\documentclass{article}

\usepackage{arxiv}

\usepackage[utf8]{inputenc} % allow utf-8 input
\usepackage[T1]{fontenc}    % use 8-bit T1 fonts
\usepackage{hyperref}       % hyperlinks
\usepackage{url}            % simple URL typesetting
\usepackage{booktabs}       % professional-quality tables
\usepackage{amsfonts}       % blackboard math symbols
\usepackage{nicefrac}       % compact symbols for 1/2, etc.
\usepackage{microtype}      % microtypography
\usepackage{lipsum}
\usepackage{graphicx}
\usepackage{enumitem}
\usepackage{ragged2e}

\usepackage{xcolor}
\hypersetup{
    colorlinks,
    linkcolor={black!50!black},
    citecolor={black!50!black},
    urlcolor={black!80!black}
}

\usepackage{graphicx}
\graphicspath{ {./images/} }

\usepackage[
backend=biber,
style=apa
]{biblatex}
\graphicspath{ {./images/} }

\title{LatinCy: Synthetic trained pipelines for Latin NLP\thanks{With annotation contributions from Nora Bernhardt (ner), Tim Geelhaar (tagger, morphologizer, parser, ner), Vincent Koch (ner).}
}

\author{
 Patrick J. Burns \\
  Institute for the Study of the Ancient World\\
  New York University\\
  \texttt{pjb311@nyu.edu}
}

\begin{document}
\maketitle
\begin{abstract}
This paper introduces LatinCy, a set of trained general purpose Latin-language “core” pipelines for use with the spaCy natural language processing framework. The models are trained on a large amount of available Latin data, including all five of the Latin Universal Dependency treebanks, which have been preprocessed to be compatible with each other. The result is a set of general models for Latin with good performance on a number of natural language processing tasks (e.g. the top-performing model yields POS tagging, 97.41\% accuracy; lemmatization, 94.66\% accuracy; morphological tagging 92.76\% accuracy). The paper describes the model training, including its training data and parameterization, and presents the advantages to Latin-language researchers of having a spaCy model available for NLP work. 
\end{abstract}


\section{Introduction} \label{introduction}
This paper introduces LatinCy, a set of trained general purpose Latin-language “core” pipelines for use with the spaCy natural language processing framework (\cite{honnibal_spacy_2023}). These are end-to-end pipelines for taking plaintext Latin as input for basic NLP processing including sentence segmentation, word tokenization, lemmatization, part-of-speech and morphological tagging, dependency parsing, and named entity recognition (NER). Three models have so far been trained, named according to spaCy conventions: \textit{la\_core\_web\_sm}, \textit{la\_core\_web\_md}, and \textit{la\_core\_web\_lg}. To clarify, ‘la’ refers to the language code for Latin, ‘core’ refers to a pipeline that includes all of the components named above, including specifically NER; ‘web’ refers to the nature of the training data, specifically that the model is trained primarily on Universal Dependency treebanks; and ‘sm’, ‘md’, and ‘lg’ refer to the “size”—i.e., small, medium, or large—of the models, with ‘md’ and ‘lg’ models being larger because they include subword vectors that describe the vocabulary while ‘sm’ models do not.

The current default pipeline consists of the following spaCy components: ‘tagger’, ‘morphologizer’, ‘trainable\_lemmatizer’ (i.e. the EditTreeLemmatizer based on \cite{muller_joint_2015}),\footnote{\url{https://explosion.ai/blog/edit-tree-lemmatizer}.} ‘parser’, and ‘ner’.\footnote{The best “description” of the pipeline, that is the documentation of all data sources, model parameters, etc. is the \texttt{project.yml} file included in the LatinCy project at \url{https://github.com/latincy}. The general “design” of the spaCy pipeline with reference to the components named here can be found at \url{https://spacy.io/models\#design}.} Sentence segmentation is trained as part of the dependency parsing task. There are also components included for orthographical regularization (i.e. overwriting the spaCy \texttt{norm\_} attribute); this is done because the ‘trainable\_lemmatizer’ can use \texttt{norm\_} for backoff instead of \texttt{text}. Splitting of the enclitic \textit{-que} is currently handled by customizing the spaCy tokenizer such that all words that end with \textit{-que} other than a provided list of exceptions (e.g. \textit{neque}, \textit{atque}, \textit{quoque}, etc.; the full list is provided in the \texttt{que\_exceptions} list in file \texttt{scripts/functions.py}) are split from their token. The spaCy ‘tok2vec’ listener is shared by all of the components at training except for ‘ner’ which is trained with a separate ‘tok2vec’ listener.\footnote{On ‘tok2vec’, see \url{https://spacy.io/api/tok2vec}.} In addition, floret vectors (50,000 or 200,000 subword vectors with a length of 300 respectively for the ‘md’ and ‘lg’ models) are trained separately and loaded into the pipeline when the training process is initialized and then used in the further training of downstream components.

The motivation for training this pipeline is as follows: 1. these are the first end-to-end trained pipelines for Latin in the spaCy “universe,” building on contributions from the author in 2022 that made Latin (‘la’) an officially supported language in spaCy,\footnote{See \url{https://github.com/explosion/spaCy/pull/11349}.} and as such make available to Latin-language researchers the work of a large NLP framework and development community; 2. these models make use of the full array of Universal Dependencies treebank annotations available for Latin by synthesizing five separate treebanks with slightly different annotations schemes into one dataset, totaling roughly 54,000 sentences with just under 1M tagged tokens; and 3. since the components are trained and evaluated using the spaCy project infrastructure, accuracy (or precision/recall as appropriate) scores can be reported for all components, and for certain components like morphological tagging and dependency parsing, these scores can be reported at feature level, significantly aiding with error analysis.

\section{Background}
The LatinCy models enter an active environment of NLP tool and model development for Latin (\cite{berti_digital_2019}), benefiting in several important ways from the work of the community, but also distinguishing themselves through platform specificity and a synthetic approach to training data compilation. End-to-end Latin pipelines are currently available through Stanza (\cite{qi_stanza_2020}) as well as the Classical Language Toolkit (\cite{johnson_classical_2021}), which bases aspects of its pipeline on tailored wrappers of the Stanza pipeline.\footnote{At the time of writing, CLTK is developing similar component wrappers for the LatinCy models.} In both cases, a single UD treebank is selected as the basis for initial training and further inference. Other than these two platforms, as discussed in \cite{burns_building_2019}, end-to-end pipelines need to be composed on an ad hoc basis, i.e. assembled sequentially from separately developed lemmatizers, POS taggers, NER taggers, etc. The bibliography in the previous cited chapter can be consulted for component-level work; notable additions since the publication of that chapter is the publication of the LiLa Knowledge Base, including related work from the CIRCSE Research Center (cf. e.g. \cite{mambrini_lila_2020}); component benchmarking contributions as part of the EvaLatin shared tasks (\cite{sprugnoli_overview_2020,sprugnoli_overview_2022}); and the publication of a transformer-based Latin language model, Latin BERT (\cite{bamman_latin_2020}). 

\section{Data}
The following open-access datasets have been used to train different parts of the LatinCy models: 1. five Latin Universal Dependencies treebanks; 2. Wikipedia and OSCAR sentence data; 3. cc100-latin, a large corpus of filtered Latin texts from the Latin-tagged texts in CC-100 (\cite{strobel_cc100-latin_2022}),  and 4. NER datasets from the Herodotos project (\cite{erdmann_practical_2019}).

\subsection{Latin Universal Dependencies} \label{latin-ud}
There are five Latin Universal Dependencies treebanks (\cite{berti_dependency_2019}) consisting of just under 1M annotated tokens: Perseus (\cite{bamman_design_2006}), PROIEL (\cite{haug_creating_2008}), ITTB (\cite{passarotti_improvements_2010,cecchini_challenges_2018}), UDante (\cite{dellorletta_udante_2020}, and LLCT (\cite{korkiakangas_late_2021}). As described below in Section \ref{preprocessing}, the UD treebanks have been preprocessed in order to create a synthetic dataset for training the following spaCy components: tagger, morphologizer, trainable\_lemmatizer, and parser. The UD treebanks have also been used in two additional ways in the pipeline: 1. all of the plaintext sentences have been extracted from the treebanks and added to the training data for the creation of the floret vectors; and 2. a curated dataset for training the NER tagger has been created using annotated versions of the UD plaintext sentences as described below in Section \ref{training-ner}.

\subsection{Wikipedia and OSCAR sentence data}
The spaCy project described below in Section \ref{training-vectors} has been used to train the floret vectors (\cite{boyd_floret_2022}) and, in following the project “recipe” as provided, it was decided to use Wikipedia (i.e. \textit{Vicipaedia}, the Latin-language Wikipedia) and OSCAR data for the basis of this task. In order to boost the quality of this dataset with relevant materials for general Latin language modeling, I also added, as mentioned above in Section \ref{latin-ud}, all of the plaintext sentences from the UD treebanks to the vector training data.

\subsection{cc100-latin}
In order to draw on a large enough vocabulary to train the ‘lg’ floret vectors (i.e. 200,000 vectors), it is necessary to use a very large collection of Latin texts. The cc100-latin dataset consists of the Latin portion of the much larger multilingual CC-100 dataset (\cite{wenzek_ccnet_2020, conneau_unsupervised_2020}). The Latin portion of CC-100 is reported as 609M tokens; this collection has been processed by Phillip Ströbel to among other thing to remove “lorem ipsum” text, deduplicate sentences, and further normalize and preprocess the text. The resulting filtered collection consists of around 390M tokens. Further details about the compilation of this text collection and its processing are available at HuggingFace Datasets.\footnote{\url{https://huggingface.co/datasets/pstroe/cc100-latin}.}

\subsection{Herodotos Project NER datasets}
As mentioned above in Section \ref{latin-ud} and further discussed below at Section \ref{training-ner}, a dataset was compiled from the existing UD annotations for training the NER component. While large, this dataset was imbalanced with a disproportionate number of PERSON annotations. In order to boost LOC (i.e. location or geographic entities) and NORP (i.e. groups of people), I added an open NER dataset from the Herodotos Project; this dataset was converted from the \texttt{.crf} files provided in the Herodotos GitHub repository into the necessary spaCy NER format.\footnote{See here for the source of the annotated Herodotos files: \RaggedRight\url{https://github.com/Herodotos-Project/Herodotos-Project-Latin-NER-Tagger-Annotation/tree/master/Annotation\_1-1-19}.} The custom NER datasets are available as part of the LatinCy projects in the directory \texttt{assets/ner/}.

\section{Methods}
\subsection{Preprocessing} \label{preprocessing}
A primary contribution of the LatinCy models is the synthesis of all available UD treebank data. As described on the “UD for Latin” site,\footnote{\url{https://universaldependencies.org/la/}.} there are variations in annotation schemes for each of the treebanks with respect to POS labels, morphological labels, and dependency relation labels. There are also differences in orthographic conventions as well as differences in tokenization and word segmentation. The goal in aligning the five treebanks is to maximize the available annotations for training components at the expense of some information loss in the interest of creating a general (and highly generalizable) Latin model. The preprocessing decisions, that is the necessary code to reproduce all preprocessing steps, are all included in the spaCy project workflows; readers are encouraged to refer to the project files for a complete description of preprocessing workflow.\footnote{The spaCy “projects” for the LatinCy models can be found in the following repositories: ‘sm’: \url{https://github.com/diyclassics/la\_core\_web\_sm}; ‘md’: \url{https://github.com/diyclassics/la\_core\_web\_md}; ‘lg’: \url{https://github.com/diyclassics/la\_core\_web\_lg}} Important aspects of the preprocessing include:
\begin{itemize}
\item UD \texttt{.conllu} files are converted to a \texttt{.tsv} file, used in the subsequent steps, so they can be loaded into the Pandas package as a dataframe and  efficiently processed through calls to the \texttt{apply} method. This is handled by the \texttt{scripts/conllu2tsv.py} file. 
\item UD lemmas are all \textit{u-v} and \textit{i-j} normalized. In the UDante treebanks, \textit{nos} and \textit{uos} are relemmatized as necessary from \textit{ego} and \textit{tu}. This is handled by the \texttt{scripts/lemma\_norm file}.
\item Sentences that have \textit{nec} tokenized as [‘c’, ‘ne’] or \textit{neque} as [‘que’, ‘ne’] are removed from the dataset. This is handled by the \texttt{scripts/remove\_perseus\_nec.py} file.
\item The UD data for UPOS, FEATURES, and XPOS are updated as follows:
    \begin{enumerate}[label=(\alph*)]
        \item UPOS tags are remapped to a smaller, more consistent set, esp. for POS tags that have different annotation schemes in the treebanks (e.g. DET and PRON or ADJ).
        \item Morphological annotations to NOUN, VERB, ADJ, DET, and PRON are limited such that they only retain gender, number, case and person, number, tense, mood, and voice; these annotations are mapped for consistency in a named tuple.
        \item XPOS tags are similarly remapped to a smaller, more consistent tagset.
    \end{enumerate}
    This is handled by the \texttt{scripts/analyze\_feats.py} file.
\end{itemize}
As mentioned above, decisions about preprocessing have been made to make the model more generalizable (i.e., e.g., through restricted tagsets) as well as more effective (i.e. by having a larger amount of training data). Because of the nature of the already robust debate around Latin annotation decisions, it is expected that some of the preprocessing decisions will be debated and altered in future iterations of model training, especially where such revised decisions contribute to better outcomes in evaluation.

\subsection{Training}
There are three primary stages in the training process for the LatinCy models: 1. training floret subword vectors; 2. training the main spaCy dependency pipeline; and 3. training the spaCy NER component. The output of the second and third stages are combined at the end of training into a single pipeline using the spaCy \texttt{assemble} method.

\subsubsection{Training floret vectors} \label{training-vectors}
Floret vectors are trained for the ‘md’ and ‘lg’ models following the workflow in the following spaCy project: “Train floret vectors from Wikipedia and OSCAR,” \textit{mutatis mutandis}, i.e. the exchange of default language (Macedonian) in the project to Latin.\footnote{\url{https://github.com/explosion/projects/tree/v3/pipelines/floret\_wiki\_oscar\_vectors}.} No parameters have been changed in training the floret vectors. As noted above, one change that has been made is the addition of UD sentence data to the vector training data in the ‘md’ vectors and the addition of both UD and cc100-latin in the ‘lg’ vectors; the total number of sentences in the training data for this task is roughly 670,000 and 11.1M, respectively. After training, the resulting model consists of 50,000 floret subword vectors of length 300 in the ‘md’ model and 200,000 in the ‘lg’ model that are used in the training of components elsewhere in the pipeline. The vectors are made available separately for download as \textit{la\_vectors\_floret\_md} and \textit{la\_vectors\_floret\_lg}.\footnote{See Section \ref{availability} below for links to the models.}

\subsubsection{Training spaCy dependency pipeline}
Most components in the LatinCy models are trained using the following spaCy project: “Part-of-speech Tagging \& Dependency Parsing (Universal Dependencies).”\footnote{\url{https://github.com/explosion/projects/tree/v3/pipelines/tagger\_parser\_ud}.} Most parameters have been left unchanged. Some notable exceptions include:
\begin{itemize}
\item ‘tok2vec’ in the ‘md’ model uses v2 (rather than v1) of the spaCy’s “MultiHashEmbed” in order to “take into account some subword information” from the available vectors.\footnote{\url{https://spacy.io/api/architectures\#MultiHashEmbed}.}
\item ‘trainable\_lemmatizer’ uses ‘norm’ (i.e. the token attribute \texttt{norm\_}) as its backoff when a suitable lemma cannot be obtained probabilistically; also, ‘min\_tree\_freq’ is lowered to ‘2’ to make more possible endings available to the model; and ‘top\_k’ is increased to ‘3’ to increase the number of possible endings to be considered before defaulting to the backoff, a decision made with consideration of the greater morphological variation in Latin (as compared to, for example, English).
\end{itemize}

It is recommended that the reader consult the \texttt{project.yml} file in the LatinCy projects for all relevant changes in training the dependency pipelines. A few additional notes about this project. The word tokenizer has been customized to support the splitting of enclitic \textit{-que} as shown in \texttt{scripts/functions.py}; the letter sequence ‘que’ is added as a suffix (as if it were punctuation) in the Latin language spaCy \texttt{Defaults} and split from tokens, just like a comma would be split from the following sequence: \textit{cano,} $\rightarrow$ [“cano”, “,”]. 

As shown in \texttt{scripts/functions.py}, the following custom components have been added to the pipeline: 1. ‘normer’; and 2. ‘lemma-fixer’. The ‘normer’ component updates the Token attribute \texttt{norm\_} so that it is \textit{u-v} normalized; this is done because the trainable lemmatizer will use \texttt{norm\_} as a backoff when unable to determine a potential lemma and this prevents \textit{v} from being retained in the \texttt{lemma\_} attribute. The ‘lemma-fixer’ component hardcodes exceptions to the lemmatizer that were having a detrimental, but easily corrected, effect on lemmatization accuracy. Specifically, it prevented the ‘que’ lemmas introduced by the custom tokenizer (and so already tagged correctly as CCONJ) from being relemmatized incorrectly as as a form of \textit{qui}; there are a number of \textit{que} (= classical \textit{quae}) examples in the training data causing this error. It also ensures that all tokens tagged as PUNCT retain their orthographic form as \texttt{lemma\_}. Lastly, labels for the components are initialized (the \texttt{init-labels} item in the project workflow) at the start of the process to speed up training.

\subsubsection{Training spaCy NER component} \label{training-ner}
The NER component is trained separately from the dependency components using the following spaCy project: “Demo NER in a new pipeline (Named Entity Recognition).”\footnote{\url{https://github.com/explosion/projects/tree/v3/pipelines/ner\_demo}.} Parameters are largely unchanged here as well. It is worth noting that the script used by the \texttt{convert-ner} item in the project workflow has been updated to read from the \texttt{.json} files that are included in \texttt{ner/assets/}; these files have been formatted in such as way as to make them more human-readable and to align better with the default NER output from the annotation software Prodigy (which will be used for the creation of additional training data for future versions of the LatinCy models). The current version of the NER component is trained for three different entity labels, following approximately categories existing already in available English-language models: 1. PERSON (“people, including fictional”); 2. LOC, a combination of the existing English GPE (“countries, cities, states”) and LOC (“non-GPE locations, mountain ranges, bodies of water”); and 3. NORP (“nationalities or religious or political groups”).\footnote{For definitions, see Section 2.6 “Entity Names Annotation” in OntoNotes v5.0 documentation: \url{https://catalog.ldc.upenn.edu/docs/LDC2013T19/OntoNotes-Release-5.0.pdf}.}

\subsection{Evaluation}
Each component in the pipeline is evaluated as part of the spaCy project workflow. The full reporting of evaluation metrics is included in the \texttt{meta.json} file packaged with the model. I have highlighted some key metrics in Table 1.

\begin{table}
 \caption{Comparison of key evaluation metrics for the LatinCy models}
  \centering
  \begin{tabular}{llll}
    \toprule
    Type     & sm score     & md score & lg score \\
    \midrule
        sentence segmentation f-score & .922 & .931 & .934 \\
        tagger accuracy (XPOS) & .932 & .935 & .941 \\
        tagger accuracy (UPOS) & .966 & .969 & .974 \\
        morphologizer accuracy & .915 & .919 & .928 \\
        trainable\_lemmatizer accuracy & .939 & .942 & .947 \\
        parser accuracy (UAS) & .821 & .818 & .831 \\
        parser accuracy (LAS) & .764 & .757 & .776 \\
        ner f-score & .889 & .892 & .908 \\
            \bottomrule
  \end{tabular}
  \label{tab:table}
\end{table}

\subsection{Retraining}
Retraining is not strictly speaking part of the pipeline creation process, but I have added it here in its own section to stress the iterative development made possible by working within the spaCy training framework. Existing pipelines can be retrained by loading the existing pipeline into a spaCy project and resuming training from this point. Components that are not being retrained can be loaded into the training framework as “frozen” components, so that specific components can be updated with the introduction of new training data.

\section{Results}
\subsection{Sample tagger output}
Figure 1 shows the kind of annotations that are returned in a spaCy Doc using the LatinCy ‘md’ model for the following sentence in \textit{Ritchie’s Fabulae Faciles}: \textit{Haec narrantur a poetis de Perseo.} Note the lemmatization error for \textit{poetis}, i.e. \textit{poetus}* (as opposed to \textit{poeta}), an incorrect word arrived at probabilistically via the edit trees created by spaCy’s ‘trainable\_lemmatizer’.

\begin{figure}
\centering
\includegraphics[width=\textwidth]{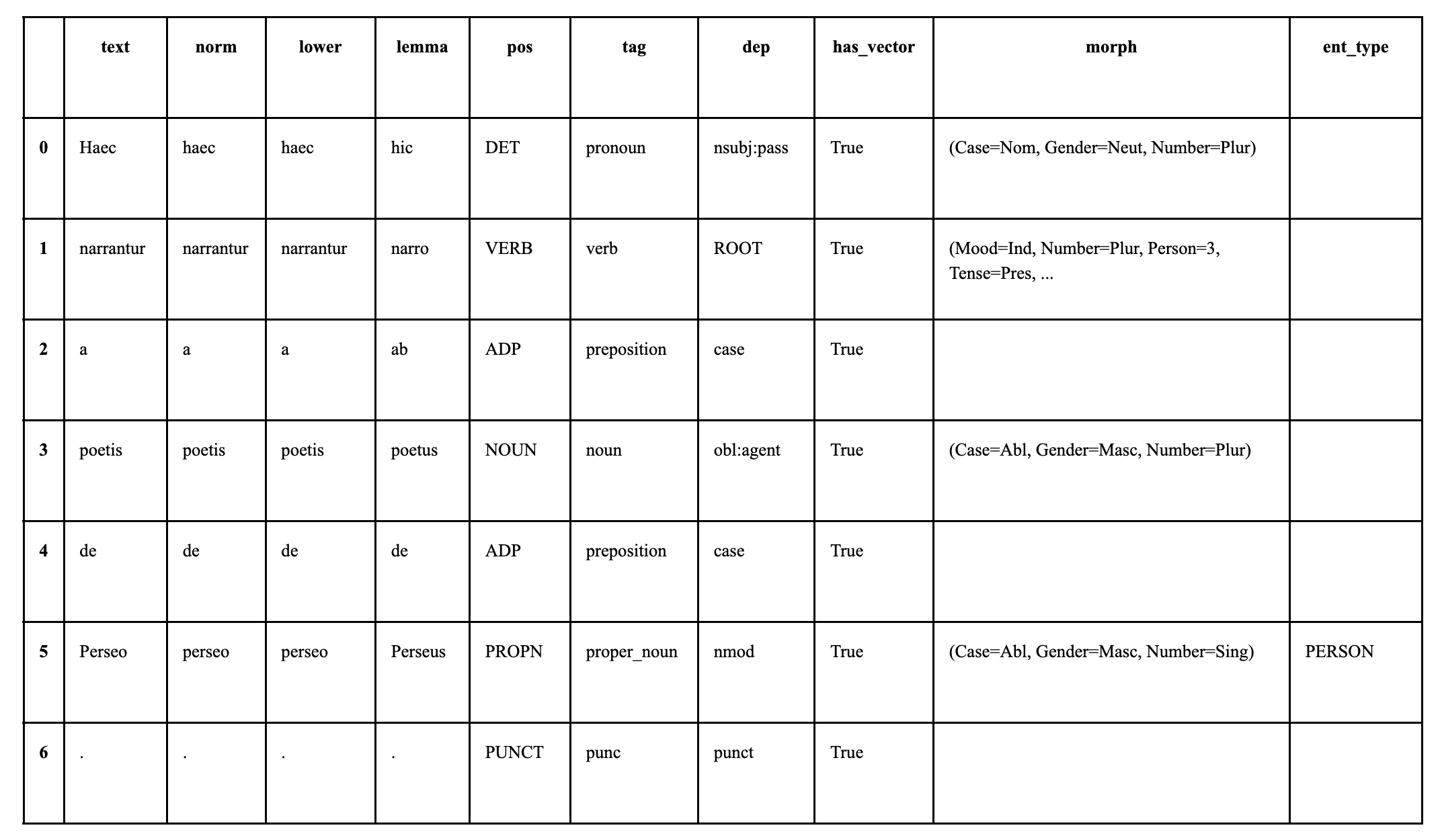}
\caption{Sample Pandas output from token annotations in the spaCy Doc as determined by the pipeline components.}
\end{figure}

\subsection{Sample dependency parser output}
Figure 2 shows an illustration using the displaCy package of the LatinCy dependency parse for the following sentence in \textit{Ritchie’s Fabulae Faciles}: \textit{Iason et Medea e Thessalia expulsi ad urbem Corinthum venerunt}.

\begin{figure}
\centering
\includegraphics[width=\textwidth]{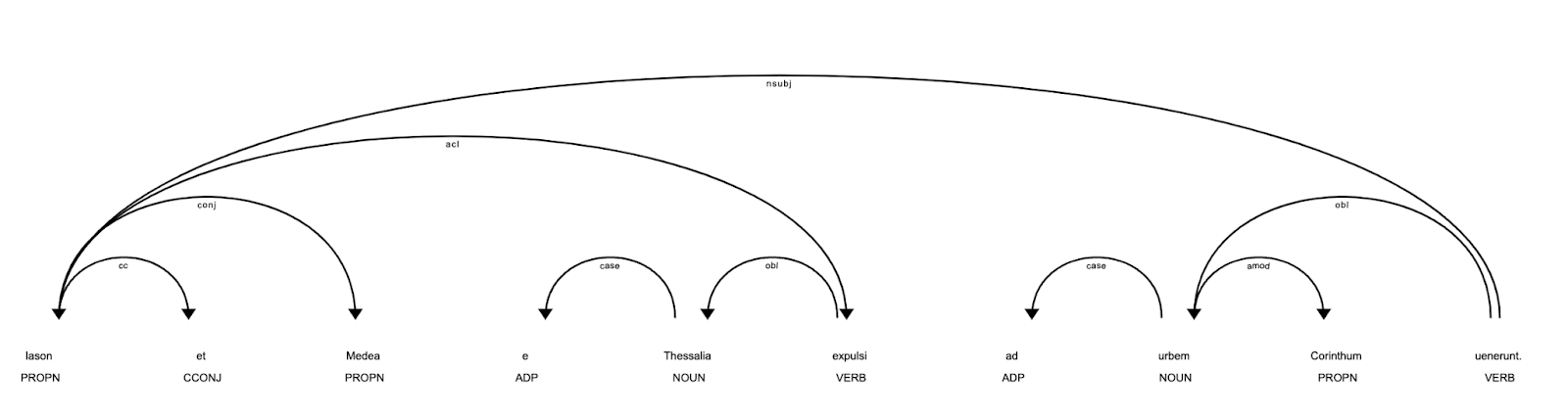}
\caption{Sample displaCy output for the dependency parser with a sample spaCy Doc.}
\end{figure}

\subsection{Sample NER output}
Figure 3 shows an illustration using the displaCy package of the named entities identified in the same sentence as above.

\begin{figure}
\centering
\includegraphics[width=\textwidth]{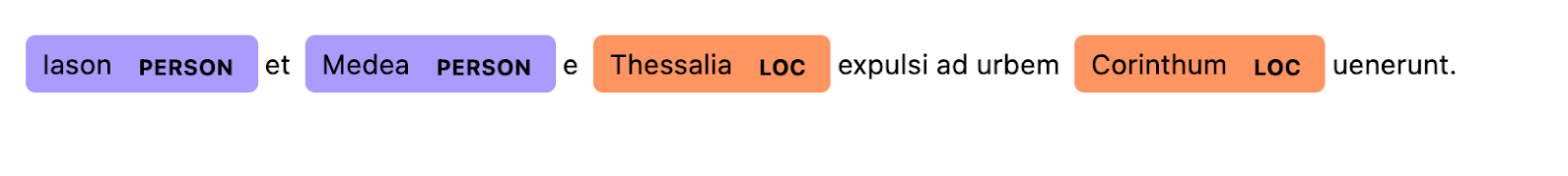}
\caption{Sample displaCy output for the NER component with a sample spaCy Doc.}
\end{figure}

\subsection{Sample vectors output}
Figure 4 shows an plot demonstrating a use case for the floret vectors included in the LatinCy ‘md’ model. Here we see plotted in two-dimensional vector space the relationship between proper names (as tagged by components earlier in the pipeline) in the complete text of \textit{Ritchie’s Fabulae Faciles}, specifically principal components (n=2) with TSNE projection.

\begin{figure}
\centering
\includegraphics[width=\textwidth]{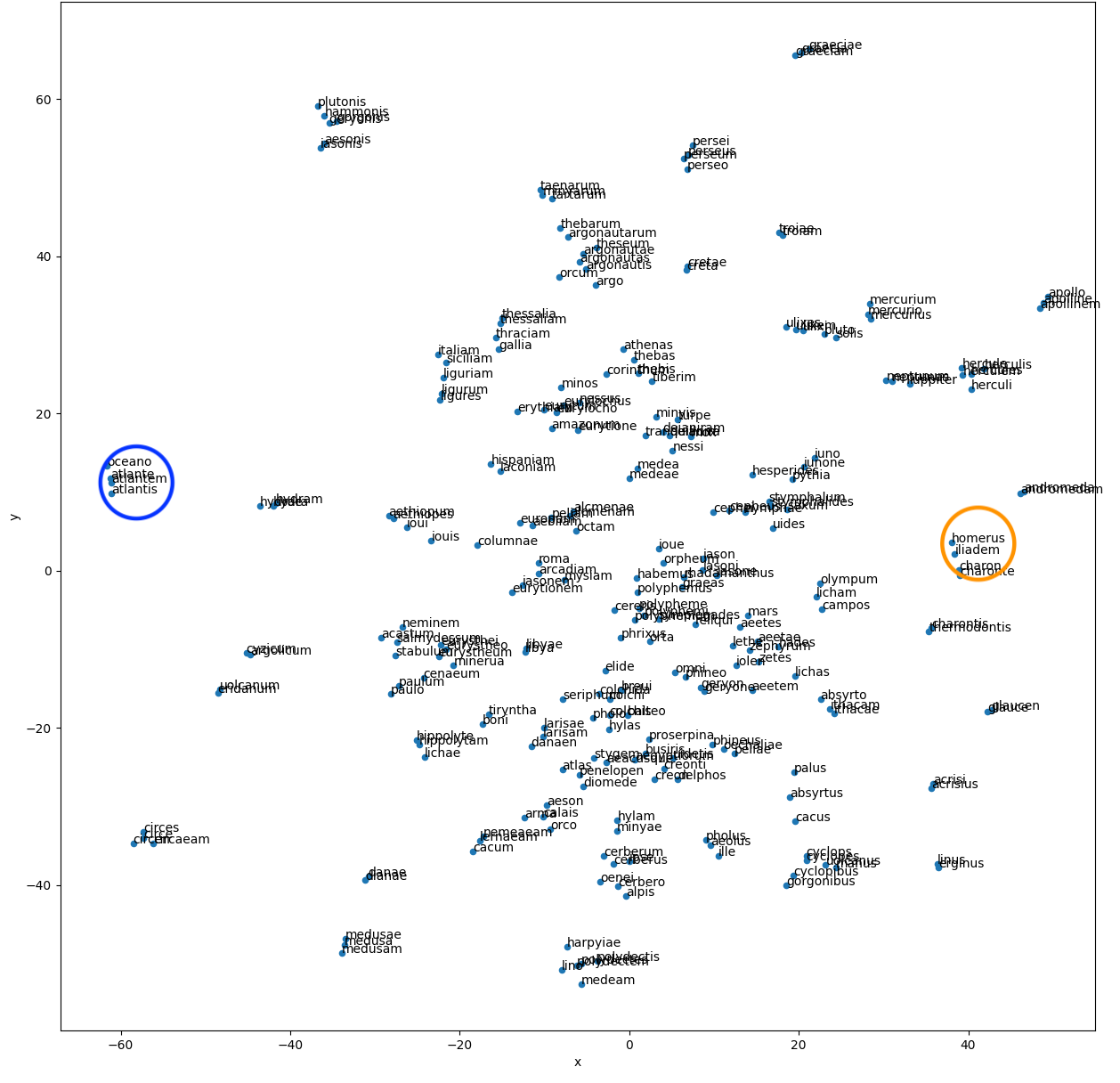}
\caption{TSNE projection of the vectors from all tokens in \textit{Ritchie’s Fabulae Faciles} that are tagged as PROPN by the LatinCy ‘md’ model.}
\end{figure}

Note the close placement, often overlap, of morphological variants of the same name (e.g. \textit{atlante, atlantem, atlantis}) circled in blue near the leftmost middle of the plot) as well as clusters of words that we would expect to appear in the same \textit{fabulae} (e.g. \textit{homerus} and \textit{iliadem} circled in orange near the rightmost middle of the plot.)

\section{Discussion}
As mentioned above in Section \ref{introduction}, one motivation for training the LatinCy pipelines is make Latin-language research compatible with a popular NLP framework with an active development community like spaCy. In this section, I present two examples of how, with the availability of these models, options in NLP are extended for Latinists.

First, with the availability of an end-to-end Latin pipeline, especially the POS tagger and dependency parser, I have been able to introduce a new “syntax iterator” to the base Latin spaCy defaults, namely noun chunks.\footnote{Latin noun chunks will be available in the spaCy v.3.6 release.} As defined in spaCy, noun chunks are “‘base noun phrases’—flat phrases that have a noun as their head. …[like] ‘the lavish green grass’ or ‘the world’s largest tech fund’.”\footnote{\url{https://spacy.io/usage/linguistic-features\#noun-chunks}.} So, for example, given the following paragraph from \textit{Ritchie’s Fabulae Faciles}—

\begin{quote}
\textit{Haec narrantur a poetis de Perseo. Perseus filius erat Iovis, \underline{maximi deorum}; \underline{avus eius} Acrisius appellabatur. Acrisius volebat Perseum \underline{nepotem suum} necare; nam propter oraculum puerum timebat. Comprehendit igitur Perseum adhuc infantem, et cum matre in \underline{arca lignea} inclusit. Tum \underline{arcam ipsam} in mare coniecit. Danae, Persei mater, magnopere territa est; tempestas enim \underline{magna mare} turbabat. Perseus autem in \underline{sinu matris} dormiebat.}
\end{quote}

—we can extract the following noun chunks (shown here as underlined text), restricting our output to chunks of more than one word: \textit{maximi deorum}, \textit{auus eius}, \textit{nepotem suum}, \textit{arca lignea}, \textit{arcam ipsam}, \textit{magna mare}, and \textit{sinu matris}. Accuracy of noun chunk extraction is a function of the accuracy of both the tagger and the parser; note, for example, that \textit{Persei mater} has been missed, in this case due to an error in the dependency parsing. This will improve with future refinement of the model. Nonetheless, we now have available a straightfoward method for extracting a large number of noun chunks from Latin text, a process that would have been difficult to perform at scale with existing Latin NLP tools. Moreover, it would be easy enough now to write additional syntax iterators for the extraction of other syntactical structures of interest to Latin researchers, such as relative clauses, \textit{cum} clauses, and so on.

Secondly, a more general way in which having an end-to-end Latin pipeline, specifically one supported by spaCy, extends NLP work in the language. I choose just one example, but considering spaCy’s widespread adoption and uses with other languages, it should be understood that there are myriad analogous possibilities. A few years back, I was researching Latin chatbot development and had become especially interested in working with the platform Rasa. We learn from the Rasa documentation that the “following components load pre-trained models that are needed if you want to use pre-trained word vectors in your pipeline,” the second of which is an end-to-end spaCy pipeline.\footnote{\url{https://rasa.com/docs/rasa/components\#spacynlp}.} The example provided in the documents links to a widely available English spaCy model (\textit{en\_core\_web\_md}) and the Rasa configuration uses this spaCy model for tokenization and feature extraction for intent classification and response classification. When I had begun looking at Rasa, the lack of a Latin spaCy model was a disincentive to continuing with that line of research. This is no longer the case with the availability of the LatinCy models and, again, there are many other NLP platforms that can now be used or used more effectively because we have a Latin model that can be used in a such “plug-and-play” fashion.

\section{Details}

\subsection{Availability} \label{availability}
The models are hosted on HuggingFace and can be installed via \texttt{pip} from this repository.\footnote{The models can be found at the following URLs: \url{https://huggingface.co/latincy/la\_core\_web\_sm}; \url{https://huggingface.co/latincy/la\_core\_web\_md}; \url{https://huggingface.co/latincy/la\_core\_web\_lg}.} The floret vectors are also available from their own HuggingFace repositories.\footnote{The vectors can be found at the following URLs: \url{https://huggingface.co/latincy/la\_vectors\_floret\_md}; \url{https://huggingface.co/latincy/la\_vectors\_floret\_lg}.} The available version at the time of writing is v.3.5.2, again following spaCy naming convention: the 3.5 refer to the version of spaCy used in training. The pipeline will be submitted for inclusion in language pipelines available for direct install through spaCy.

\subsection{Assets and rights acknowledgements}
The UD datasets are available for use under the following licenses: Perseus (CC BY-NC-SA 2.5), PROIEL (CC BY-NC-SA 3.0), ITTB (CC BY-NC-SA 3.0), UDante (CC BY-NC-SA 3.0), and LLCT (CC BY-SA 4.0). The treebanks are not republished as part of the project; these assets are downloaded as part of the spaCy project workflow and removed at the end of training. The sentences from these treebanks, in the required NER training data format, are published in the LatinCy projects as described in Section 3.2. The Herodotos Project data is available for use under a GNU Affero General Public License v3.0 license.

\section{Acknowledgments}
This work is made possible by the open licenses under which the five Latin treebanks have been released; without the efforts of the treebank project maintainers, and especially the treebank annotators, there would not be LatinCy models. This is also true for the Wikipedia, OSCAR, Herodotos Project,\footnote{\url{https://u.osu.edu/herodotos/team/}.} and cc100-latin data.\footnote{\url{https://www.cl.uzh.ch/de/people/team/compling/pstroebel.html}.} I want to acknowledge the support of the Institute for the Study of the Ancient World at New York University as well as their commitment to computational work on ancient-world data. I also thank David Bamman, Gregory Crane, Christopher Francese, William J. Mattingly (and the CLTK maintainers), and David Mimno for their early feedback on this model and the spaCy maintainers for their interest in including Latin among their language offerings.

\printbibliography

@misc{bamman_latin_2020,
	title = {Latin {BERT}: A Contextual Language Model for Classical Philology},
	url = {http://arxiv.org/abs/2009.10053},
	author = {Bamman, David and Burns, Patrick J.},
	date = {2020},
	langid = {english},
	eprinttype = {arxiv},
	eprint = {2009.10053}
}

@inproceedings{bamman_design_2006,
	location = {Prague},
	title = {The Design and Use of a {L}atin Dependency Treebank},
	pages = {67--78},
	booktitle = {Proceedings of the Fifth Workshop on Treebanks and Linguistic Theories ({TLT}2006)},
	publisher = {Ú{FAL} {MFF} {UK}},
	author = {Bamman, David and Crane, Gregory},
	date = {2006}
}

@book{berti_digital_2019,
	location = {Berlin},
	title = {Digital Classical Philology: {A}ncient {G}reek and {L}atin in the Digital Revolution},
	url = {https://www.degruyter.com/view/product/502894},
	publisher = {De Gruyter},
	author = {Berti, Monica},
	date = {2019}
}

@online{boyd_floret_2022,
	title = {floret: Lightweight, Robust Word Vectors},
	url = {https://explosion.ai/blog/floret-vectors},
	titleaddon = {explosion.ai {B}log \& {N}ews},
	author = {Boyd, Adriane and Warmerdam, Vincent D.},
	date = {2022},
	langid = {english}
}

@incollection{burns_building_2019,
	location = {Berlin},
	title = {Building a Text Analysis Pipeline for Classical Languages},
	url = {https://www.degruyter.com/view/books/9783110599572/9783110599572-010/9783110599572-010.xml},
	pages = {159--176},
	booktitle = {Digital Classical Philology: {A}ncient {G}reek and {L}atin in the Digital Revolution},
	publisher = {De Gruyter},
	author = {Burns, Patrick J.},
	date = {2019}
}

@incollection{berti_dependency_2019,
	title = {The Dependency Treebanks for {A}ncient {G}reek and {L}atin},
	isbn = {978-3-11-059957-2},
	url = {https://www.degruyter.com/view/book/9783110599572/10.1515/9783110599572-016.xml},
	pages = {279--298},
	booktitle = {Digital Classical Philology: {A}ncient {G}reek and {L}atin in the Digital Revolution},
	publisher = {De Gruyter},
	author = {Celano, Giuseppe G. A.},
	editor = {Berti, Monica},
	date = {2019},
	langid = {english},
	doi = {10.1515/9783110599572-016}
}

@inproceedings{cecchini_challenges_2018,
	title = {Challenges in converting the {I}ndex {T}homisticus Treebank into {U}niversal {D}ependencies},
	pages = {27--36},
	booktitle = {Proceedings of the Second Workshop on Universal Dependencies ({UDW} 2018},
	author = {Cecchini, F.M. and Passarotti, M. and Marongiu, P. and Zeman, D.},
	date = {2018},
	langid = {english}
}

@incollection{dellorletta_udante_2020,
	title = {{UDante}: First Steps Towards the {U}niversal {D}ependencies Treebank of {D}ante’s {L}atin Works},
	isbn = {9791280136336},
	url = {http://books.openedition.org/aaccademia/8653},
	pages = {99--105},
	booktitle = {Proceedings of the Seventh Italian Conference on Computational Linguistics {CLiC}-it 2020},
	publisher = {Accademia University Press},
	author = {Cecchini, F.M. and Sprugnoli, Rachele and Moretti, Giovanni and Passarotti, Marco},
	editor = {Dell'Orletta, Felice and Monti, Johanna and Tamburini, Fabio},
	date = {2020},
	langid = {english},
	doi = {10.4000/books.aaccademia.8653}
}

@inproceedings{conneau_unsupervised_2020,
	location = {Online},
	title = {Unsupervised Cross-lingual Representation Learning at Scale},
	url = {https://www.aclweb.org/anthology/2020.acl-main.747},
	doi = {10.18653/v1/2020.acl-main.747},
	eventtitle = {Proceedings of the 58th Annual Meeting of the Association for Computational Linguistics},
	pages = {8440--8451},
	booktitle = {Proceedings of the 58th Annual Meeting of the Association for Computational Linguistics},
	publisher = {Association for Computational Linguistics},
	author = {Conneau, Alexis and Khandelwal, Kartikay and Goyal, Naman and Chaudhary, Vishrav and Wenzek, Guillaume and Guzmán, Francisco and Grave, Edouard and Ott, Myle and Zettlemoyer, Luke and Stoyanov, Veselin},
	date = {2020},
	langid = {english}
}

@inproceedings{erdmann_practical_2019,
	location = {Minneapolis, Minnesota},
	title = {Practical, Efficient, and Customizable Active Learning for Named Entity Recognition in the Digital Humanities},
	booktitle = {Proceedings of North American Association of Computational Linguistics ({NAACL} 2019)},
	author = {Erdmann, Alexander and Wrisley, David Joseph and Allen, Benjamin and Brown, Christopher and Bodénès, Sophie Cohen and Elsner, Micha and Feng, Yukun and Joseph, Brian and Joyeaux-Prunel, Béatrice and Marneffe, Marie-Catherine},
	date = {2019},
	langid = {english}
}

@inproceedings{haug_creating_2008,
	title = {Creating a parallel treebank of the old {Indo-European} Bible translations},
	pages = {27--34},
	booktitle = {Proceedings of the Second Workshop on Language Technology for Cultural Heritage Data ({LaTeCH} 2008},
	author = {Haug, D.T. and Jøhndal, M.},
	date = {2008},
	langid = {english}
}

@software{honnibal_spacy_2023,
	title = {{spaCy}: Industrial-strength Natural Language Processing in {P}ython},
	url = {https://spacy.io/},
	version = {v. 3.5.2},
	author = {Honnibal, Matthew and Montani, Ines},
	date = {2023}
}

@inproceedings{johnson_classical_2021,
	title = {The {Classical Language Toolkit}: An {NLP} Framework for Pre-Modern Languages},
	rights = {All rights reserved},
	url = {https://aclanthology.org/2021.acl-demo.3},
	doi = {10.18653/v1/2021.acl-demo.3},
	pages = {20--29},
	booktitle = {Proceedings of the 59th Annual Meeting of the Association for Computational Linguistics and the 11th International Joint Conference on Natural Language Processing: System Demonstrations},
	publisher = {Association for Computational Linguistics},
	author = {Johnson, Kyle P. and Burns, Patrick J. and Stewart, John and Cook, Todd and Besnier, Clément and Mattingly, William J. B.},
	date = {2021}
}

@article{korkiakangas_late_2021,
	title = {{L}ate {L}atin {C}harter {T}reebank: contents and annotation},
	volume = {16},
	issn = {1749-5032},
	url = {https://www.euppublishing.com/doi/abs/10.3366/cor.2021.0217},
	doi = {10.3366/cor.2021.0217},
	pages = {191--203},
	number = {2},
	journaltitle = {Corpora},
	author = {Korkiakangas, Timo},
	date = {2021}
}

@article{mambrini_lila_2020,
	title = {{LiLa}: {L}inking {L}atin. Risorse linguistiche per il latino nel Semantic Web},
	volume = {4},
	url = {https://umanisticadigitale.unibo.it/article/view/9975},
	doi = {10.6092/issn.2532-8816/9975},
	number = {8},
	journaltitle = {Umanistica Digitale},
	author = {Mambrini, Francesco and Cecchini, Flavio Massimiliano and Franzini, Greta and Litta, Eleonora and Passarotti, Marco Carlo and Ruffolo, Paolo},
	date = {2020-05-18},
	langid = {italian}
}

@inproceedings{muller_joint_2015,
	location = {Lisbon, Portugal},
	title = {Joint Lemmatization and Morphological Tagging with {L}emming},
	url = {http://aclweb.org/anthology/D15-1272},
	doi = {10.18653/v1/D15-1272},
	eventtitle = {Proceedings of the 2015 Conference on Empirical Methods in Natural Language Processing},
	pages = {2268--2274},
	booktitle = {Proceedings of the 2015 Conference on Empirical Methods in Natural Language Processing},
	publisher = {Association for Computational Linguistics},
	author = {Müller, Thomas and Cotterell, Ryan and Fraser, Alexander and Schütze, Hinrich},
	date = {2015},
	langid = {english}
}

@inproceedings{passarotti_improvements_2010,
	location = {Valletta, Malta},
	title = {Improvements in Parsing the {I}ndex {T}homisticus Treebank. Revision, Combination and a Feature Model for Medieval {L}atin},
	url = {http://www.lrec-conf.org/proceedings/lrec2010/pdf/178_Paper.pdf},
	eventtitle = {{LREC} 2010},
	booktitle = {Proceedings of the Seventh International Conference on Language Resources and Evaluation ({LREC}'10)},
	publisher = {European Language Resources Association ({ELRA})},
	author = {Passarotti, Marco and Dell'Orletta, Felice},
	date = {2010-05}
}

@article{qi_stanza_2020,
	title = {Stanza: A {P}ython Natural Language Processing Toolkit for Many Human Languages},
	url = {http://arxiv.org/abs/2003.07082},
	journaltitle = {{arXiv}:2003.07082 [cs]},
	author = {Qi, Peng and Zhang, Yuhao and Zhang, Yuhui and Bolton, Jason and Manning, Christopher D.},
	date = {2020-04-23},
	langid = {english}
}

@inproceedings{sprugnoli_overview_2020,
	location = {Marseille},
	title = {Overview of the {EvaLatin} 2020 Evaluation Campaign},
	eventtitle = {Language Resources and Evaluation Conference ({LREC} 2020)},
	pages = {105--110},
	booktitle = {Proceedings of 1st Workshop on Language Technologies for Historical and Ancient Languages},
	author = {Sprugnoli, Rachele and Passarotti, Marco and Cecchini, Flavio M and Pellegrini, Matteo},
	date = {2020-05-11},
	langid = {english}
}

@article{sprugnoli_overview_2022,
	title = {Overview of the {EvaLatin} 2022 Evaluation Campaign},
	pages = {6},
	author = {Sprugnoli, Rachele and Passarotti, Marco and Cecchini, Flavio Massimiliano and Fantoli, Margherita and Moretti, Giovanni},
	date = {2022},
	langid = {english},
	booktitle = {Proceedings of Second Workshop on Language Technologies for Historical and Ancient Languages},
    pages = {183--188},
    url = {https://aclanthology.org/2022.lt4hala-1.29}
}

@misc{strobel_cc100-latin_2022,
	title = {{c}c100-latin},
	url = {https://huggingface.co/datasets/pstroe/cc100-latin},
	author = {Ströbel, Phillip},
	date = {2022}
}

@inproceedings{wenzek_ccnet_2020,
	title = {{CCNet}: Extracting High Quality Monolingual Datasets from Web Crawl Data},
	pages = {4003--4012},
	booktitle = {Proceedings of the 12th Conference on Language Resources and Evaluation ({LREC} 2020)},
	author = {Wenzek, Guillaume and Lachaux, Marie-Anne and Conneau, Alexis and Chaudhary, Vishrav and Guzmán, Francisco and Joulin, Armand and Grave, Edouard},
	date = {2020},
	langid = {english}
}

\end{document}